\documentclass{article}
\PassOptionsToPackage{numbers,compress}{natbib}

\usepackage[main, final]{neurips_2026}

\usepackage[utf8]{inputenc} 
\usepackage[T1]{fontenc}    
\usepackage{hyperref}       
\usepackage{url}            
\usepackage{float}          
\usepackage{booktabs}       
\usepackage{amsfonts}       
\usepackage{nicefrac}       
\usepackage{microtype}      
\usepackage{xcolor}         
\usepackage{graphicx}       
\usepackage{amsmath}        
\usepackage{multirow}       
\usepackage{wrapfig}
\usepackage{caption}
\usepackage{algorithm}
\usepackage{algorithmic}

\title{Scene-Adaptive VLA: Efficient Autonomous Driving via Dynamic Layer Routing}

\author{%
  \textbf{Yujia Yang} \quad
  \textbf{Quan Yuan}\thanks{Corresponding author} \quad
  \textbf{Jiawei Fan} \quad
  \textbf{Yang Li} \\
  \textbf{Tiange Fu} \quad
  \textbf{Xiaoyuan Fu} \quad
  \textbf{Guiyang Luo} \quad
  \textbf{Jinglin Li} \\[0.75ex]
  State Key Laboratory of Networking and Switching Technology, \\
  Beijing University of Posts and Telecommunications \\
  {\footnotesize\texttt{\{yangyj2022,yuanquan,fjw208,leeyang866,tgfu,fuxiaoyuan,luoguiyang,jlli\}@bupt.edu.cn}}
}

\begin{document}

\maketitle

\begin{abstract}
 Vision-Language-Action (VLA) models offer a promising paradigm for autonomous driving. However, the massive parameter scale of VLAs strains onboard computational resources during real-time inference. This directly increases energy consumption and limits vehicle battery range, making it critical to minimize computational overhead without compromising driving capability and explainability. Current research on efficient VLA-based driving focuses on sequential adaptation (\textit{e.g.}, Chain-of-Thought and token adjustment), leaving structural adaptation underexplored despite parameter redundancy across dynamic driving scenes. To address this gap, we propose Scene-Adaptive VLA, an efficient framework leveraging dynamic layer routing to adjust active parameters based on real-time scenes. Inspired by the temporal continuity of scenes and model predictive uncertainty, our approach characterizes model-perceived scene complexity via frame-level temporal scene variations and decision deviations. To uncover latent scene-to-parameter correspondences, we encode this characterization into scene-aware tokens, alongside two specialized queries. A budget-conditioned layer routing mechanism evaluates these refined queries to allocate a budget that determines the active parameter ratio and subsequently selects appropriate layer combinations. The entire framework is optimized end-to-end. Extensive evaluations on the Bench2Drive benchmark show that when adapted to a state-of-the-art VLA model, our approach significantly reduces inference-time computational overhead by 53.5\% with negligible degradation in driving and language capabilities. Code is available at \url{https:github.com/yang-luvu/Scene-Adaptive-VLA}.
\end{abstract}

\section{Introduction}
\label{sec:introduction}
Built on Vision-Language Models (VLMs), Vision-Language-Action (VLA) models have emerged as a dominant paradigm in autonomous driving, unifying visual perception, linguistic reasoning, and action prediction within a cohesive framework \cite{xu2025drivegpt4,sima2024drivelm,renz2025simlingo,fu2025orion}.
However, the massive parameter scale of VLA models creates severe onboard computational resource pressure for real-time inference, directly increasing energy consumption and reducing vehicle battery range.
Consequently, minimizing computational overhead without compromising driving capability and explainability has become a paramount objective in VLA-based autonomous driving research \cite{zhou2025autovla,yang2025drivemoe}.
\begin{figure}[h]
    \centering
    \includegraphics[width=\linewidth]{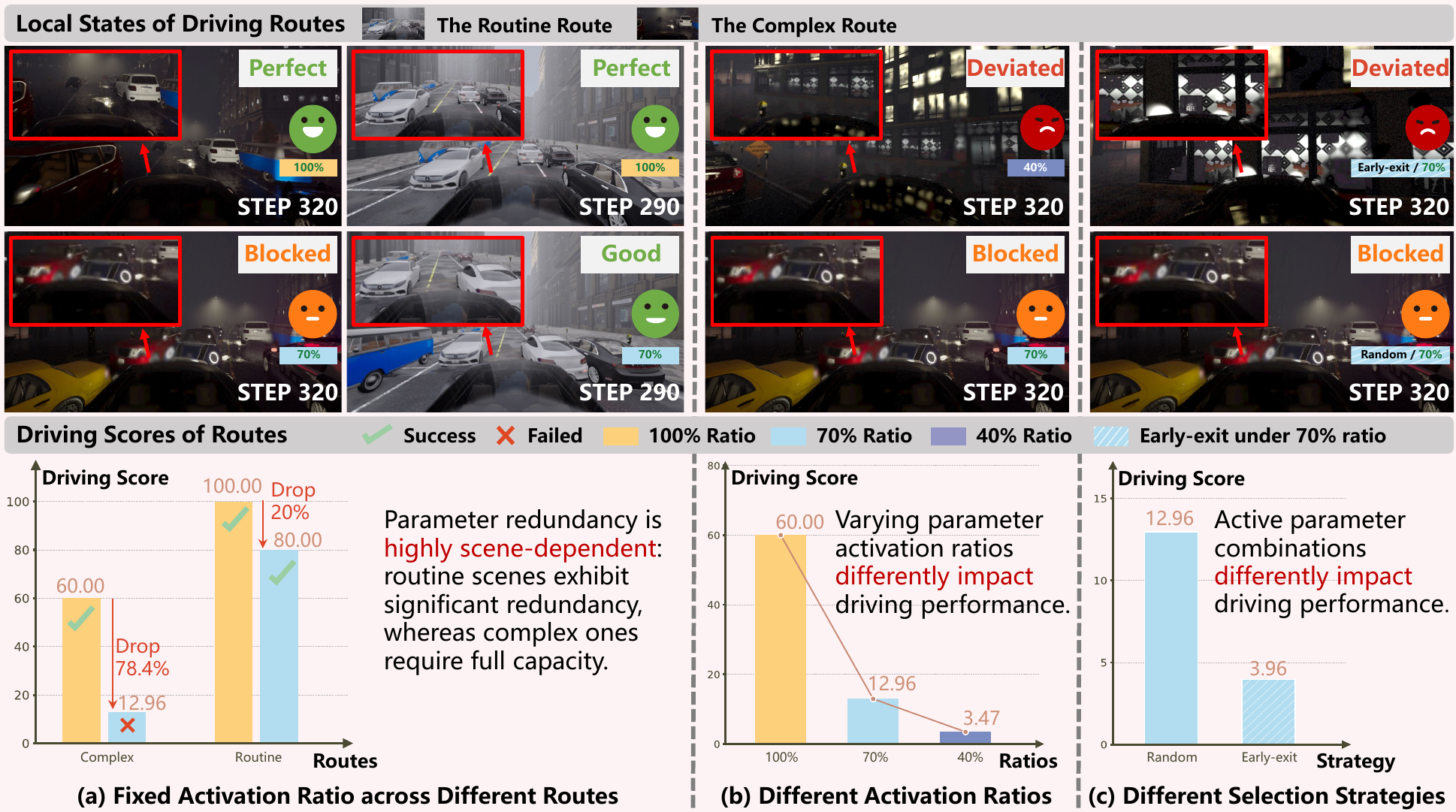}
    \caption{Driving performance under varying capacities of a state-of-the-art model \cite{renz2025simlingo} on Bench2Drive. (a) Fixed Ratio Across Routes: The model with 70\% active parameters completely fails on routes featuring more complex driving scenes, yet completes routes with only a 20\% score drop on routes dominated by routine scenes. (b) Activation Ratios (Complex Route): Reducing parameters severely degrades performance, with lower ratios causing critical local failures (e.g., lane deviation at 40\%, blockage at 70\%). (c) Layer Selection Strategies (Complex Route): At 70\% activation, an early-exit strategy significantly underperforms random layer dropping, proving that active parameter combinations are critical.
    }
    \label{fig:introduction}
    \vspace{-0.6cm}
\end{figure}
Currently, efforts to improve inference efficiency in VLMs can be broadly categorized into two directions: \textit{sequential adaptation}, which dynamically adjusts the number of input or output tokens based on the task \cite{shang2025llava,liu2026global,zhang2025continue,zhang2025adaptthink}, and \textit{structural adaptation}, which selectively activates model parameters during inference to adapt to the task \cite{xu2025learning}.
In the domain of VLA-based autonomous driving, existing efficient inference strategies predominantly focus on sequential adaptation, such as adjusting the length of Chain-of-Thought (CoT) reasoning \cite{zhou2025autovla} or modulating input token counts based on the scene \cite{yang2025drivemoe}. Conversely, structural adaptation remains largely underexplored.

As illustrated in Fig.~\ref{fig:introduction}(a), the ratio of parameters required for successful driving varies significantly across scenes of varying complexity. Structural adaptation thus holds substantial potential to enable dynamic, task-aware adjustment of active parameters, thereby facilitating efficient autonomous driving.
Furthermore, varying the active parameter ratio and combination yields distinctly different performance within identical scenes as shown in Fig.~\ref{fig:introduction}(b) and (c). Therefore, we aim to design a VLA framework that dynamically determines the optimal parameter ratio and combination for each driving frame. However, establishing this scene-to-parameter mapping is non-trivial due to the considerable diversity of driving scenes and the abstract nature of model capabilities. Specifically, determining the \textit{scene-to-parameter ratio} is challenging because perceiving scene complexity and translating it into a measure of required model capacity is a highly implicit process. Furthermore, optimizing the \textit{scene-to-parameter combination} is complex. Accurately selecting optimal active parameters for a specific driving scene is complicated by the entangled black-box nature of neural networks and the high variance of driving scenarios.

To address these challenges, we propose Scene-Adaptive VLA, an efficient VLA-based autonomous driving framework leveraging dynamic layer routing. Inspired by the temporal continuity of driving scenes and model predictive uncertainty theory \cite{ovadia2019can,lakshminarayanan2017simple,cao2024pasco}, Scene-Adaptive VLA characterizes model-perceived scene complexity via frame-level temporal scene variations and decision deviations.
To uncover latent scene-to-parameter correspondences, we formulate scene-aware tokens that encode the aforementioned scene characterization, alongside two specialized queries: a budget token for determining parameter ratios and a routing token for selecting layer combinations. After processing these tokens through shallow layers of the Large Language Model (LLM) to capture their complex interdependencies, we introduce a budget-conditioned layer routing mechanism. This mechanism evaluates the refined queries to first determine the active parameter ratio and subsequently route inputs to the optimal layer combinations. Finally, the entire framework is optimized via two-stage fine-tuning supervised by both driving performance and resource consumption, effectively establishing an efficient scene-to-parameter mapping. Theoretically, our framework can be integrated into prevailing VLA architectures that employ dense backbones. To validate its effectiveness under hardware constraints, we adapt our framework to SimLingo \cite{renz2025simlingo}, which achieves optimal performance with minimal parameter scale in VLA-based driving models. The main contributions of this work are summarized as follows:

1. A novel formulation for the characterization of model-perceived scene complexity. Leveraging temporal continuity of driving scenes and model predictive uncertainty theory, it effectively characterizes scene complexity via temporal frame-level driving scene variations and decision deviations, encoding them into scene-aware tokens.

2. A lightweight, budget-conditioned routing mechanism for efficient dynamic parameter activation. Utilizing two queries alongside the extracted scene-aware tokens and the LLM's shallow layers, this mechanism dynamically allocates a real-time inference budget to determine the active parameter ratio and selects the optimal parameter combination via layer routing, all optimized end-to-end.

3. Exceptional closed-loop performance on the Bench2Drive benchmark. When adapted to the state-of-the-art (SOTA) VLA-based driving model, our framework achieves a 53.5\% reduction in inference-time computational overhead with negligible degradation in driving proficiency and language capabilities, as demonstrated by extensive evaluations on the Bench2Drive benchmark and language tasks.

\section{Related Work}
\textbf{VLA-based End-to-end Autonomous Driving.}
End-to-end autonomous driving has witnessed rapid advancements in recent years. These frameworks integrate perception and planning within unified architectures to enhance driving efficacy \cite{hu2023planning,jiang2023vad}. To address complex long-tail scenarios, recent approaches integrate VLMs either for language-centric scenario understanding \cite{park2024vlaad, sima2024drivelm} or for guiding downstream driving models \cite{pan2024vlp, jiang2024senna}. However, these decoupled paradigms lack direct end-to-end optimization grounded in continuous actions. Consequently, VLA architectures have emerged to directly couple VLMs with action generation. While Drivegpt4-v2 \cite{xu2025drivegpt4} maps multimodal inputs to control signals and EMMA \cite{hwang2024emma} unifies tasks via plain text representations, a critical modality gap remains. To bridge this gap between semantic reasoning and numerical control, ORION \cite{fu2025orion} incorporates a generative planner alongside a query-based temporal module, enabling unified optimization for scenario reasoning and trajectory prediction. Concurrently, SimLingo \cite{renz2025simlingo} enforces strict language-action alignment via an auxiliary task trained on synthetic instruction-action pairs, yielding superior performance in closed-loop evaluation.

\textbf{Efficient VLA-based Model Inference in Autonomous Driving.}
Despite their formidable capabilities, the immense parameter scale of VLA models imposes a prohibitive computational burden on vehicular hardware. To address this, recent studies have explored efficient inference strategies. For instance, AutoVLA \cite{zhou2025autovla} reduces reasoning overhead by introducing a fast-and-slow thinking mechanism that dynamically adjusts CoT lengths via Group Relative Policy Optimization \cite{shao2024deepseekmath} (GRPO). Similarly, DriveMoE \cite{yang2025drivemoe} alleviates the cost of long vision sequences by employing a Mixture-of-Experts architecture with on-demand vision token filtering. While effective, these methods largely overlook active parameter redundancy during inference.

\textbf{Structural Adaptation in VLMs.}
Structural adaptation optimizes the performance-efficiency trade-off by dynamically allocating active parameters based on task demands. For Vision Transformers and LLMs, this is typically achieved by selectively dropping computational modules or employing adaptive early-exit mechanisms \cite{zhu2025ea, meng2020ar, rotem2023finding}. Extending these paradigms to VLMs, AdaLLaVA \cite{xu2025learning} introduces a framework that selectively disables entire Transformer layers using binary switches under an inference latency constraint. For structural adaptation in VLA-based autonomous driving, it is essential to determine the optimal parameter activation ratio and enable flexible parameter combinations conditioned on driving scenes. Thus, we adopt a similar paradigm to achieve active-ratio-aware layer selection for VLA models.

\section{Scene-Adaptive VLA}
\label{sec:Scene-Adaptive-VLA}
\begin{figure}[h]
  \centering
  \includegraphics[width=\linewidth]{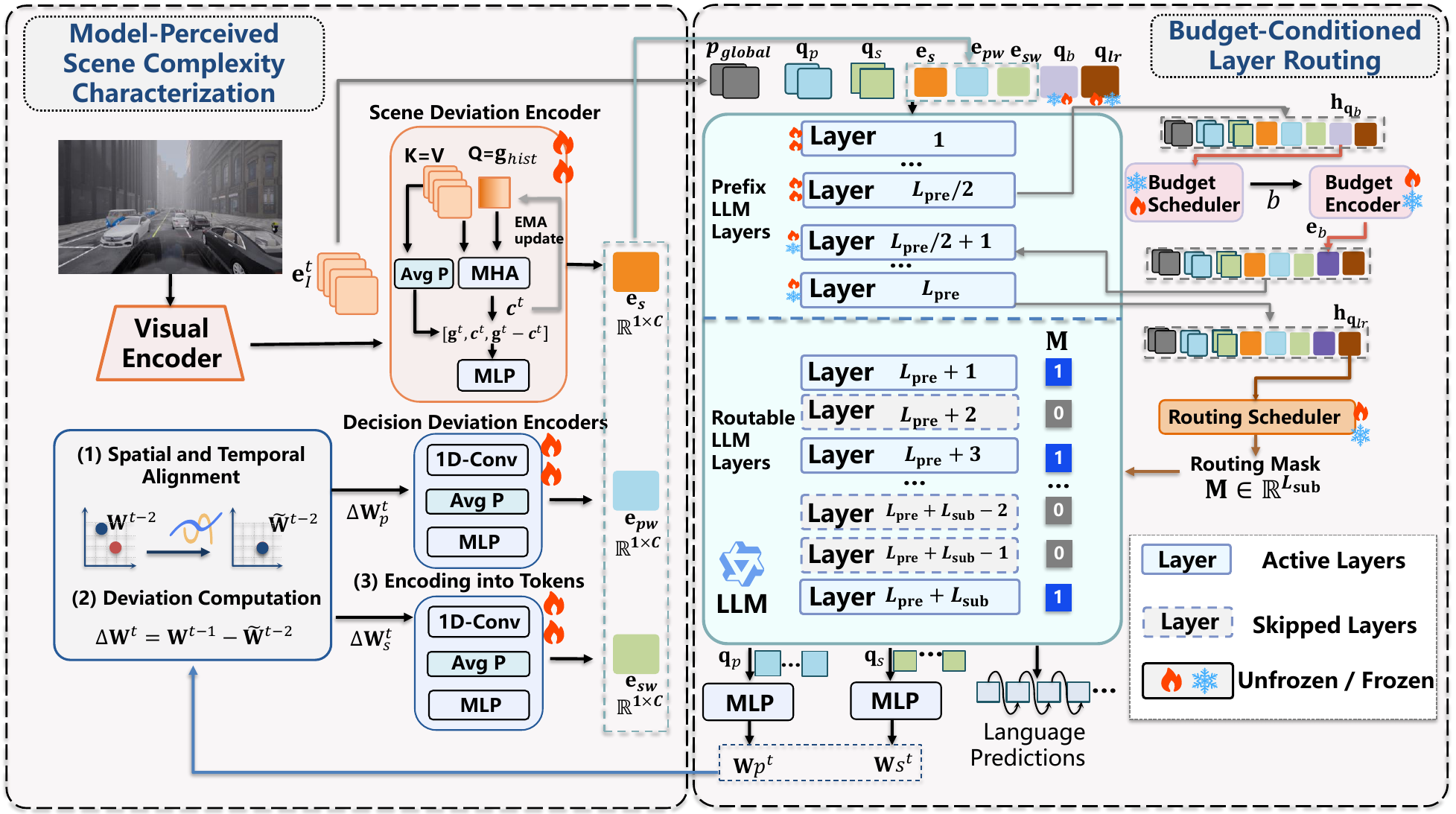}
  \caption{Overview of the Scene-Adaptive VLA, featuring Model-Perceived Scene Complexity Characterization and Budget-Conditioned Layer Routing. Adjacent icons denote paramter status (left/top: Stage 1; right/bottom: Stage 2). \textbf{Left}: The characterization module encodes temporal visual variations and inter-frame decision deviations into scene-aware tokens ($\mathbf{e}_s$, $\mathbf{e}_{pw}$, $\mathbf{e}_{sw}$). \textbf{Right}: The LLM prefix layers process these tokens alongside budget ($\mathbf{q}_b$) and routing ($\mathbf{q}_{lr}$) queries. A budget scheduler allocates a real-time computational budget $b$ based on intermediate $\mathbf{q}_b$ features from the middle prefix layers. At the final prefix layer, a routing scheduler utilizes $\mathbf{q}_{lr}$ to generate a binary mask $\mathbf{M}$, dynamically activating specific layers and bypassing others via residual connections to yield final predictions.}
  \label{fig:method}
  \vspace{-0.3cm}
\end{figure}

To validate our method under hardware constraints, we adapt it to SimLingo \cite{renz2025simlingo}, a VLA driving model that achieves optimal performance with minimal parameters. As illustrated in Fig.~\ref{fig:method}, our approach enables scene-adaptive parameter activation through two core components. First, Model-Perceived Complexity Characterization encodes scene complexity into scene-aware tokens by quantifying temporal visual variations and inter-frame decision deviations. Second, Budget-Conditioned Layer Routing dynamically allocates a real-time computational budget to dictate the active parameter ratio and layer combinations. These decisions are governed by budget and layer routing queries, which are processed in the shallow layers of the LLM alongside scene-aware tokens to capture scene-parameter correlations. The training proceeds in two stages. Stage 1 employs SFT with randomized budgets to initialize scene and routing representations. Stage 2 leverages GRPO \cite{shao2024deepseekmath} for end-to-end optimization, utilizing a composite reward derived from driving proficiency and computational overhead.

\subsection{Framework}

\textbf{Base VLA Model}.
The backbone is InternVL2-1B \cite{gao2024mini}, comprising the InternViT-300M vision encoder and the Qwen2-0.5B-Instruct \cite{yang2407qwen2} LLM.
To ensure lightweight and efficient execution, we partition the language model into $L_{\text{pre}}$ shallow prefix layers and $L_{\text{sub}}$ subsequent routable layers. We augment this architecture with two compact multi-layer perceptron (MLP) schedulers. Operating on intermediate features extracted by the prefix layers, these schedulers leverage the infused scene-aware tokens and specialized queries to dynamically determine the routing plan for the remaining layers without introducing substantial computational overhead.

\textbf{Input Representation}.
Following SimLingo, the driving-specific inputs comprise the visual scene tokens $\mathbf{e}_I \in \mathbb{R}^{N_c \times C}$, the task prompt $p_{\text{task}}$, navigation tokens $\mathbf{e}_{\text{nav}}$, ego-vehicle speed $v$, path waypoint queries $\mathbf{q}_p \in \mathbb{R}^{N_p \times C}$ and speed waypoint queries $\mathbf{q}_s \in \mathbb{R}^{N_s \times C}$, where $N_c$, $N_p$, and $N_s$ denote the respective token counts, and $C$ represents the LLM hidden dimension. The elements form a global prompt $p_{\text{global}}$: \texttt{<image features $\mathbf{e}_I$>\textbackslash n Current speed: <$v$> m/s. Command: <nav. features $\mathbf{e}_{\text{nav}}$>. <task prompt $p_{\text{task}}$>}.
To facilitate scene-adaptive layer routing, we introduce five additional tokens in $\mathbb{R}^{1 \times C}$. The scene-aware inputs comprise a scene deviation token $\mathbf{e}_s$ alongside decision deviation tokens for path ($\mathbf{e}_{pw}$) and speed ($\mathbf{e}_{sw}$), with their construction detailed in Section 3.2. Specialized queries consist of the budget query $\mathbf{q}_b$ and the layer routing query $\mathbf{q}_{lr}$, which govern budget allocation and layer routing, respectively. The complete sequence fed into the model is concatenated in order as $I_{\text{llm}} = [p_{\text{global}}, \mathbf{q}_p, \mathbf{q}_s, \mathbf{e}_s, \mathbf{e}_{pw}, \mathbf{e}_{sw}, \mathbf{q}_b, \mathbf{q}_{lr}]$.

\subsection{Model-Perceived Scene Complexity Characterization}
To realize scene-adaptive parameter activation, quantifying complexity from the model's perspective is essential. Drawing on the principle that prediction variance with similar inputs indicates uncertainty \cite{lakshminarayanan2017simple,ovadia2019can}, we characterize complexity via two complementary signals. The first signal is temporal scene-level deviations that capture dynamic visual variations. The second signal is inter-frame decision deviations, reflecting model uncertainty through decision discrepancies across adjacent frames. Together, these signals provide a scene-complexity-aware representation to inform downstream budget and layer routing decisions.

\textbf{Scene Temporal Deviation.} 
Building upon the visual scene tokens $\mathbf{e}_I^t$ extracted by the visual encoder, we obtain a global scene representation at time $t$, $\mathbf{g}^t$, via mean pooling:
$\mathbf{g}^t = \frac{1}{N_c}\sum_{i=1}^{N_c} \mathbf{e}_I^{t,i},$ where $\mathbf{e}_I^{t,i} \in \mathbb{R}^C$.
To capture temporal variations, we maintain a historical scene representation $\mathbf{g}_{\text{hist}}$, updated across frames via an exponential moving average (EMA). A multi-head cross-attention (MHA) mechanism explicitly models temporal evolution by:
\begin{align}
\label{eq:cross_attn}
\mathbf{c}^t = \text{MHA}(\text{query}=\mathbf{g}_{\text{hist}}, \text{key}=\mathbf{e}_I^t, \text{value}=\mathbf{e}_I^t),
\mathbf{g}_{\text{hist}} = \alpha \mathbf{c}^t + (1 - \alpha) \mathbf{g}_{\text{hist}},
\end{align}
where $\alpha \in [0, 1]$ controls the update rate and $\mathbf{c}^t$ captures deviations from the expected visual state. The scene deviation token $\mathbf{e}_s$ is then computed by concatenating features along the channel dimension:
\begin{align}
\label{eq:es}
\mathbf{e}_s = \text{MLP}_s([\mathbf{g}^t, \mathbf{c}^t, \mathbf{g}^t - \mathbf{c}^t]) \in \mathbb{R}^{1 \times C}.
\end{align}
This design encodes both the absolute scene content and its historical deviation into a single compact token.

\textbf{Decision Deviation.}
To maintain an efficient single-forward-pass inference paradigm, we evaluate the decision deviation of the current timestep $t$ by leveraging cached outputs from the two preceding timesteps, $t-1$ and $t-2$. This is feasible due to the substantial temporal overlap between consecutive long-horizon predicted trajectories.
Let $\mathbf{W}_p^{t-1}$ and $\mathbf{W}_p^{t-2}$ denote the predicted cumulative-sum path waypoints from the previous two frames. After adjusting $\mathbf{W}_p^{t-2}$ for ego-motion via relative translation and yaw, we apply linear interpolation for temporal alignment to obtain the aligned past trajectory $\tilde{\mathbf{W}}_p^{t-2}$. The per-waypoint path deviation is $\Delta \mathbf{W}_p = \mathbf{W}_p^{t-1} - \tilde{\mathbf{W}}_p^{t-2}$. This sequence is condensed into a unified path deviation token $\mathbf{e}_{pw}$ via:
\begin{align}
\label{eq:path_dev_token}
\mathbf{e}_{pw} = \text{MLP}_{pw}(\text{AvgPooling}(\text{Conv1D}_{pw}(\Delta \mathbf{W}_p))) \in \mathbb{R}^{1 \times C}.
\end{align}
Similarly, for the predicted speed waypoints $\mathbf{W}_s \in \mathbb{R}^{N_s \times 2}$, we apply identical spatial-temporal alignment to obtain $\tilde{\mathbf{W}}_s^{t-2}$. By measuring the Euclidean distance between consecutive waypoints, we derive the sequential speed profiles $\mathbf{v}^{t-1}$ and $\tilde{\mathbf{v}}^{t-2} \in \mathbb{R}^{(N_s-1) \times 1}$. The speed deviation $\Delta \mathbf{v} = \mathbf{v}^{t-1} - \tilde{\mathbf{v}}^{t-2}$ is encoded into $\mathbf{e}_{sw} \in \mathbb{R}^{1 \times C}$ using an analogous network:
\begin{align}
\label{eq:speed_dev_token}
\mathbf{e}_{sw} = \text{MLP}_{sw}(\text{AvgPooling}(\text{Conv1D}_{sw}(\Delta \mathbf{v}))) \in \mathbb{R}^{1 \times C}.
\end{align}
By concatenating these scene-aware tokens directly into the input sequence, the shallow prefix layers can deeply fuse them with the intrinsic scene features, establishing the foundation for the budget-conditioned layer routing detailed in Section 3.3.

\subsection{Budget-Conditioned Layer Routing Mechanism}
To uncover latent scene-to-parameter correspondences, we introduce a budget-conditioned layer routing mechanism. First, a budget allocation module determines the active parameter ratio via the budget query. Subsequently, a routing module selects optimal layer combinations guided by the layer routing query under the allocated budget. The framework is then optimized through two-stage fine-tuning that balances driving performance and computational overhead to establish an efficient scene-to-parameter mapping.

\textbf{Scene-Adaptive Layer Routing Inference.}
As illustrated in Fig.~\ref{fig:method}, midway through the $L_{\text{pre}}$ prefix layers, the budget scheduler extracts the hidden feature of the budget query token, $\mathbf{h}_{q_b}$, to select a discrete budget action $a\in\{0,...,K-1\}$.
This selected action is subsequently scaled to determine the assigned budget $b=\max(\frac{1}{K} \cdot (a+1),\, b_{\min})$. The lower bound, $b_{\min}$, guarantees a foundational reasoning capacity for the model and is defined as $b_{\min} = \max(0.2, L_{\text{pre}}/L_{\text{llm}})$, where $L_{\text{llm}} = L_{\text{pre}} + L_{\text{sub}}$.
The budget $b$ is then encoded into $\mathbf{e}_b$ via sinusoidal embeddings and a budget encoder $f_{\text{enc}}$, replacing the initial $\mathbf{q}_b$ token. At the prefix boundary, the routing scheduler $f_{\text{routing}}$ processes the concatenated layer routing query $\mathbf{q}_{lr}$ and $\mathbf{e}_b$ to generate routing logits $\boldsymbol{\pi}_{\text{layer}} \in \mathbb{R}^{L_{\text{sub}}}$ for the $L_{\text{sub}}$ subsequent layers. 

To enforce the budget constraint, $b$ is quantized into an active sub-layer count $k = \max(0, \lfloor L_{\text{llm}} \cdot b \rfloor - L_{\text{pre}})$. Then we obtain the binary routing mask $\mathbf{M} \in \{0,1\}^{L_{\text{sub}}}$ via Gumbel-Softmax top-$k$ sampling:
\begin{align}
\label{eq:gumbel_routing}
\mathbf{M} = \text{GumbelSoftmax}_\tau(\boldsymbol{\pi}_{\text{layer}})_{\text{top-}k}, \text{s.t.} \; \mathbf{M}_j \in \{0,1\}, \; \sum_{j=1}^{L_{\text{sub}}} \mathbf{M}_j = k.
\end{align}
The straight-through estimator ensures valid gradient flow during training. The mask $\mathbf{M}$ acts as a binary gate for the remaining decoder layers, efficiently bypassing unselected layers via residual connections to decode language tokens and waypoints. To optimize this architecture effectively, we propose a two-stage training pipeline.

\textbf{Stage 1 Scene-aware Representation and Layer Routing Learning.}
Stage 1 initializes scene-aware representations and budget-aware routing via SFT. With the budget scheduler frozen, budgets are uniformly sampled from $b \sim \mathcal{U}[b_{\min}, 1]$. We jointly optimize the layer routing query $\mathbf{q}_{lr}$, the scene-aware encoders, the routing scheduler $f_{\text{routing}}$, the budget encoder $f_{\text{enc}}$, and the LLM prefix layers (tuned via LoRA \cite{hu2022lora}). The primary supervision minimizes the Smooth-L1 waypoint loss $\mathcal{L}_{\text{wp}}$ and the auto-regressive language loss $\mathcal{L}_{\text{lang}}$, following the formulation in SimLingo:
\begin{align}
\label{eq:loss_wp_lang}
\mathcal{L}_{\text{wp}} &= \text{SmoothL1}(\mathbf{W}_p, \mathbf{W}_p^{\text{gt}}) + \text{SmoothL1}(\mathbf{W}_s, \mathbf{W}_s^{\text{gt}}),
\mathcal{L}_{\text{lang}} = -\sum_{t=1}^{T} \log P(l_t \mid l_{<t}, I_{\text{llm}}).
\end{align}
Here $\mathbf{W}_p^{\text{gt}}$, $\mathbf{W}_s^{\text{gt}}$, and $l_t$ denote ground-truth path waypoints, speed waypoints, and language tokens, respectively. To prevent degenerate routing behaviors, we introduce two auxiliary regularizers. The diversity loss $\mathcal{L}_{\text{div}}$ encourages distinct routing patterns for dissimilar scenes. The balance loss $\mathcal{L}_{\text{bal}}$ aligns the empirical per-layer activation rate $\bar{a}_j$ with the target budget ratio $\bar{b}$:
\begin{align}
\label{eq:loss_div_bal}
\mathcal{L}_{\text{div}} = \frac{1}{B^2}\sum_{i,j} \big[\text{sim}(\mathbf{M}_i, \mathbf{M}_j) - \text{sim}(\mathbf{z}_i, \mathbf{z}_j) - m_{\text{d}}\big]_+^2, \;
\mathcal{L}_{\text{bal}} = \frac{1}{L_{\text{sub}}}\sum_{j} \big[|\bar{a}_j - \bar{b}| - \varepsilon_{\text{b}}\big]_+^2,
\end{align}
where $[\cdot]_+$ is the ReLU function, $\text{sim}(\cdot, \cdot)$ is cosine similarity, and $m_{\text{d}}$ and $\varepsilon_{\text{b}}$ are margin hyperparameters. $\mathbf{z}_i$ is the max-pooled scene feature. $\bar{a}_j = \frac{1}{B}\sum_i \mathbf{M}_{i,j}$ represents the mean activation rate of sub-layer $j$ over batch $B$. The overall Stage 1 objective is the weighted sum:
$
\mathcal{L}_{\text{stage1}} = \lambda_{\text{wp}} \mathcal{L}_{\text{wp}} + \lambda_{\text{lang}} \mathcal{L}_{\text{lang}} + \lambda_{\text{div}}\mathcal{L}_{\text{div}} + \lambda_{\text{bal}}\mathcal{L}_{\text{bal}}.
$

\textbf{Stage 2 Reinforcement Fine-Tuning for Budget Allocation.}
To explicitly learn an optimal budget allocation policy, we apply GRPO \cite{shao2024deepseekmath}. We unfreeze the budget scheduler $f_{\text{budget}}$ and the budget query $\mathbf{q}_b$ while freezing the routing scheduler $f_{\text{routing}}$, the budget encoder $f_{\text{enc}}$, the layer routing query $\mathbf{q}_{lr}$, and the latter half of the prefix layers. The policy $\pi_\theta(a \mid \mathbf{h}_{q_b}) = \text{Categorical}\big(\text{softmax}(f_{\text{budget}}(\mathbf{h}_{q_b}))\big)$ maps the intermediate representation to a distribution over discrete budget actions. For each sample $i$, we draw $G$ actions $\{a_{i,g}\}_{g=1}^{G}$ from the old policy $\pi_{\text{old}}$. The reward $r_{i,g}$ balances driving quality improvements against budget parsimony:
\begin{align}
\label{eq:reward_advantage}
r_{i,g} = w_d \tanh\left(\mathcal{L}_{\text{full},i} - \mathcal{L}_{i,g}\right) - w_b \frac{b(a_{i,g}) - b_{\min}}{1 - b_{\min}},
A_{i,g} = \frac{r_{i,g} - \mu_i}{\sigma_i + \varepsilon_a} + w_a r_{i,g},
\end{align}
where $\mathcal{L}_{\text{full},i}$ and $\mathcal{L}_{i,g}$ denote the full-budget reference loss and the loss under action $a_{i,g}$. Weights $w_d$, $w_b$, and $w_a$ control the driving-quality, budget-parsimony, and absolute driving-performance terms, respectively. $\mu_i$ and $\sigma_i$ are the mean and standard deviation of rewards across the $G$ groups, and $\varepsilon_a$ ensures numerical stability. The policy is updated via a clipped surrogate objective with KL divergence regularization:
\begin{align}
\label{eq:loss_policy}
\mathcal{L}_{\text{policy}} = -\mathbb{E}_{i,g}\Big[\min\big(\rho_{i,g}(\theta) A_{i,g}, \text{clip}(\rho_{i,g}(\theta), 1-\varepsilon_c, 1+\varepsilon_c)A_{i,g}\big)\Big] + \beta \text{KL}(\pi_\theta \parallel \pi_{\text{old}}),
\end{align}
where $\rho_{i,g}(\theta) = \frac{\pi_\theta(a_{i,g} \mid \mathbf{h}_{q_b}^{(i)})}{\pi_{\text{old}}(a_{i,g} \mid \mathbf{h}_{q_b}^{(i)})}$ is the importance sampling ratio. $\theta$ and $\theta_{\text{old}}$ denote the current and old policy parameters, $\varepsilon_c$ and $\beta$ are hyperparameters controlling the clipping range and the weight of the KL.
To prevent catastrophic deviation in the unfrozen parameters of the prefix layers' LoRA and scene-aware encoders, we introduce an anchor loss:
$
\mathcal{L}_{\text{anchor}} = \frac{1}{|\boldsymbol{\phi}|}\sum_j (\phi_j - \bar{\phi}_j)^2.
$
Here, $\boldsymbol{\phi}$ represents the vector of currently trainable parameters, and $\bar{\boldsymbol{\phi}}$ denotes their corresponding frozen values obtained from Stage 1. The total Stage 2 objective is therefore:
$
\mathcal{L}_{\text{stage2}} = \mathcal{L}_{\text{policy}} + \beta_{\text{anchor}} \mathcal{L}_{\text{anchor}}.
$

\section{Experiments}
\label{sec:experiments}
\subsection{Benchmarks and Metrics}
\textbf{Inference Computational Overhead.}
To demonstrate the efficiency of our approach, we report the estimated per-decision floating-point operations (FLOPs) alongside measured runtime metrics, including per-decision latency and peak memory. We compute FLOPs for the LLM component only, which dominates the computation of these VLA models. All runtime metrics are measured on Bench2Drive under identical settings using an RTX 4090. Language generation is enabled for all methods except DriveMoE, which does not support language output. Since all compared methods use dense decoder-only LLM backbones, we estimate LLM computation as $\mathrm{FLOPs}_{\mathrm{LLM}} \approx 2P(N_{\mathrm{in}}+N_{\mathrm{out}})$ \cite{kaplan2020scaling,shoeybi2019megatron}, where \(P\) is average number of active LLM parameters, and \(N_{\mathrm{in}}\) and \(N_{\mathrm{out}}\) denote the average number of input and output tokens, respectively. Detailed token configurations are provided in Appendix~\ref{app:computational_overhead}. For baselines with dynamic computation, we report their lowest-compute settings, i.e., AutoVLA in fast-thinking mode and DriveMoE with single-view input.
\begin{table}[H]
  \centering
  \caption{LLM computation overhead per driving decision.}
  \label{tab:FLOPS}
  \setlength{\tabcolsep}{3pt}
  \resizebox{\linewidth}{!}{%
  \begin{tabular}{@{}l|ccccccc@{}}
    \toprule
    \multicolumn{1}{c|}{Method} & Avg. $P$(B) & $N_{\text{in}}$ & $N_{\text{out}}$
    & TFLOPs $\downarrow$ & Latency (ms) $\downarrow$
    & Mem. (GB) $\downarrow$ & Language \\
    \midrule
    DriveMoE\cite{yang2025drivemoe} & 3.00 & 527 & 1 & 3.17 & \underline{319.54} & 14.85 & $\times$ \\
    AutoVLA\cite{zhou2025autovla} & 3.00 & 1736 & 18 & 10.52 & 1322.85 & 7.18 & $\checkmark$ \\
    ORION\cite{fu2025orion} & 7.00 & 538 & 110 & 9.07 & 5440.70 & 17.83 & $\checkmark$ \\
    ORION+SA & 3.68 & 538 & 110 & 4.77 & 2971.60 & 17.83 & $\checkmark$ \\
    SimLingo\cite{renz2025simlingo} & 0.50 & 569 & 22 & \underline{0.59} & 410.79 & \underline{2.03} & $\checkmark$ \\
    SimLingo+SA & 0.23 & 569--576 & 22
    & \textbf{0.27} & \textbf{214.01} & \textbf{2.01} & $\checkmark$ \\
    \bottomrule
  \end{tabular}%
  }
\end{table}
\textbf{Bench2Drive.}
We evaluate closed-loop performance on the Bench2Drive benchmark \cite{jia2024bench2drive} based on the CARLA simulator, comprising 220 diverse 150-meter routes. We report four primary official metrics: Driving Score, Success Rate, Efficiency, and Comfortness. Success rate measures the proportion of routes successfully completed within the time limit, while driving score incorporates both route completion status and violation penalties. Efficiency and Comfortness evaluate the autonomous driving system's speed performance and passenger comfort during navigation, respectively.

\textbf{Driving Linguistic Capabilities Test.} To assess the impact of our approach on the base VLA model's language abilities, we evaluate its performance on the linguistic benchmarks introduced by SimLingo using SPICE \cite{anderson2016spice} as the metric. Specifically, we utilize the DriveLM-hard visual question answering (VQA) and the driving Commentary validation set. This dataset comprises 4,355 VQA test samples and 4,678 driving commentary samples, which primarily focus on interpreting the current driving frame. Consequently, the VQA subset effectively tests general scene comprehension, while the Commentary subset systematically measures the model's capacity to articulate driving decisions and their underlying rationales in natural language.

\subsection{Implementation Details}
\textbf{Model Setting.} We integrate Scene-Adaptive VLA into SimLingo, denoted as SimLingo+SA. We train our model using the dataset provided by SimLingo, jointly optimizing for driving and language tasks on two NVIDIA RTX 4090 GPUs. Based on the LLM architecture, we set the number of prefix layers to $L_{\text{pre}} = 4$ and the routable sub-layers to $L_{\text{sub}} = 20$. During training, LoRA is applied to the prefix layers to enhance efficiency. In Stage 1, we use a learning rate of 3e-5 and a batch size of 6, training for one epoch (approximately 90k steps). The loss weights for waypoints and language modeling are set to $\lambda_{\text{wp}} = \lambda_{\text{lang}} = 1.0$, with diversity and balance regularization weights $\lambda_{\text{div}} = 0.05$ and $\lambda_{\text{bal}} = 0.01$, respectively. In Stage 2, the learning rate is adjusted to 1e-5 with a batch size of 4, training for roughly 40k steps. GRPO is configured with a group size of $G=4$ and performs three policy updates per step; the weights for driving performance, budget, and absolute driving performance are $w_{d} = 1.0, w_{b} =0.05, w_{a}=0.3$.

\begin{table}[t]
  \centering
  \caption{Comparison with SOTA driving methods on the closed-loop benchmark.}
  \label{tab:bench2drive}
  \begin{tabular*}{\linewidth}{@{\extracolsep{\fill}}l|ccccc@{}}
    \toprule
    \multicolumn{1}{c|}{Method} & DS$\uparrow$ & SR($\%$) $\uparrow$ & Efficiency$\uparrow$ & Comfortness$\uparrow$ & TFLOPs$\downarrow$  \\
    \midrule
      UniAD-Base\cite{hu2023planning}    & 45.81  & 16.36 & 129.21 & 43.58   & 3.42\\
      VAD\cite{jiang2023vad}    & 42.35 & 15.00 & 157.94 & 46.01   & 1.14\\
      TCP-traj\cite{wu2022trajectory}    & 59.90 & 30.00 & 76.54 & 18.08   & \textbf{0.04}\\
      AutoVLA\cite{zhou2025autovla} & 78.84 & 57.73 & 146.93 & 39.33  & 10.52 \\
      DriveMoE\cite{yang2025drivemoe}    & 74.22 & 48.64 & 175.96 & 15.31  & 3.17 \\
      ORION\cite{fu2025orion} & 77.74 & 54.62 & 151.48 & 17.38 & 9.07 \\
      ORION+SA@100\% & 77.45 & 53.63 & 153.03 & 21.38 & 9.07\\
      ORION+SA@65\% & 75.07 & 49.54 & 152.47 & 22.88 & 5.90\\
      ORION+SA & 75.88 & 50.45 & 150.09 & 25.29 & 4.77 \\
      SimLingo\cite{renz2025simlingo}   & 85.07 & \textbf{67.27} & 259.23 & 33.67  & 0.59 \\
      SimLingo+SA    & 84.87 & \underline{65.45} & \textbf{266.19} & \textbf{54.36} & 0.27 \\
      SimLingo+SA@100\%    & \textbf{86.83} & \textbf{67.27} & \underline{262.13}  & 44.79  & 0.59\\
      SimLingo+SA@70\%    & \underline{85.44} & 64.09 & 259.06  & 53.19  & 0.41\\
      SimLingo+SA@40\%    & 78.20 & 59.09 & 262.05  & \underline{53.20}  & \underline{0.24}\\
    \bottomrule
  \end{tabular*}
\end{table}
\subsection{Main Results}
\textbf{Inference Computational Overhead.}
We evaluate the average LLM computational overhead per driving decision across VLA models using FLOPs and measured runtime metrics. In all tables within this section, the best results are highlighted in bold, and the second-best results are underlined. As Tab.~\ref{tab:FLOPS} shows, our framework reduces SimLingo's FLOPs by 53.5\% and latency by 47.9\% via scene-adaptive dynamic parameter activation. For comparison, we further evaluate SimLingo+SA against efficient VLA baselines, including DriveMoE and AutoVLA. Despite disabling language generation, DriveMoE is still slower and uses 7.4$\times$ more peak memory, while AutoVLA incurs substantially higher latency even in fast-thinking mode. In addition, our lightweight framework design contributes significantly to this efficiency: it introduces auxiliary modules comprising a mere 14M parameters, and is specifically designed to exclusively process the 5 additional input tokens through the initial prefix layers. This incurs virtually no extra computational overhead due to the shallow depth of these layers.

\vspace{1em} 
\noindent 
\begin{minipage}[c]{0.48\textwidth} 
\textbf{Closed-loop Driving Performance.}
We evaluate the closed-loop driving performance of SimLingo+SA. The results are averaged over three random seeds. As demonstrated in Tab.~\ref{tab:bench2drive}, our method significantly reduces the computational overhead of the original SimLingo framework by 53.5\% while incurring only a marginal performance degradation, specifically a 0.2\% drop in DS and a 1.82\% drop in SR. 
\end{minipage}\hfill 
\begin{minipage}[c]{0.48\textwidth} 
  \centering
  \setlength{\tabcolsep}{3.5pt}
  \captionof{table}{Language ability evaluation.}
  \label{tab:language_ability}
  \begin{tabular}{@{}l|ccc@{}}
    \toprule
    \multicolumn{1}{c|}{Method} & Budget & VQA$\uparrow$ & Commentary$\uparrow$ \\
    \midrule
    InternVL2-2B & - & 44.40 & 9.04 \\
    SimLingo & - & \textbf{56.77} & \textbf{38.04}    \\
    SimLingo+SA & 100\% & \underline{54.91} & \underline{35.37}   \\
    SimLingo+SA & 70\% & 53.55 & 35.28   \\
    \bottomrule
  \end{tabular}
\end{minipage}
This demonstrates that our approach can dynamically select the appropriate parameter ratio and combination conditioned on the driving scene, thereby substantially reducing the runtime resource usage of VLA models.
Furthermore, by effectively capturing inter-frame dynamics and decision deviations, our model surpasses SimLingo in Efficiency and notably enhances Comfortness, achieving the best results.
Additionally, we evaluate the driving performance of our model under fixed budgets, denoted as SimLingo+SA@$b$. Under a full budget, our framework improves all other driving metrics while maintaining the Success Rate, demonstrating that our approach enables dynamic scene-adaptive parameter adjustment while improving the foundational model's performance ceiling. The multi-ability results are shown in Appendix \ref{app:multi-ability}.

\textbf{Linguistic Capabilities.}
To verify the impact of our method on linguistic reasoning capabilities, we evaluate our model on the driving VQA dataset introduced by SimLingo. As shown in Tab.~\ref{tab:language_ability}, although our framework incurs a slight performance degradation compared to the original model, specifically a 1.86\% drop in DriveLM-VQA and a 2.67\% drop in Commentary under full budget, it effectively preserves the explainability of driving scenes and decisions, consistently outperforming InternVL2-2B under different budgets.

\begin{figure}[h]
  \centering
  \includegraphics[width=\linewidth]{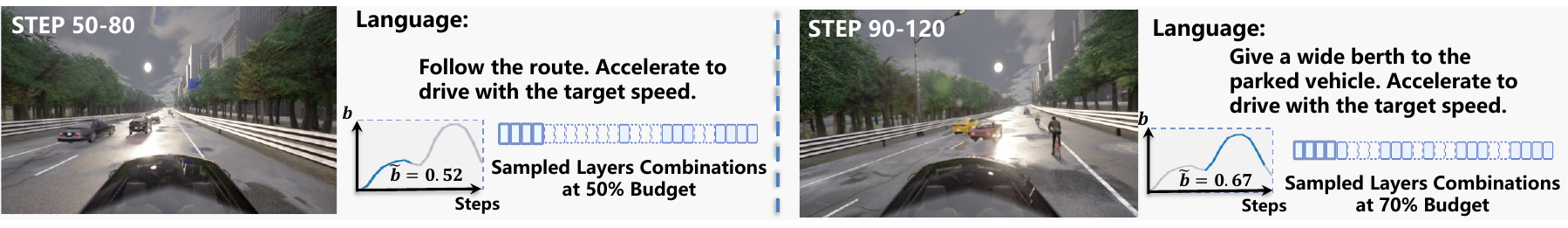} 
  \captionof{figure}{Visualization of scene-adaptive parameter activation. Left: driving scenes at selected steps. Right: corresponding budget and sampled layer combinations.}
  \label{fig:qualitative}
\end{figure}

\textbf{Qualitative Analysis.}
To illustrate the dynamic parameter activation capability of our Scene-Adaptive VLA, we visualize selected driving frames from Route 3 of the Bench2Drive benchmark as shown in Fig.~\ref{fig:qualitative}. Scenes between driving steps 50 and 80 are relatively routine, requiring ego-vehicle merely to follow the lane; thus, an average of only 52\% of the parameters is activated. As the vehicle progressively approaches a dense traffic area, the allocated inference budget gradually increases. Subsequently, during steps 90 to 120, a slow-moving cyclist appears ahead, necessitating a lane change to maintain a safe distance. Consequently, a higher proportion of parameters is dynamically allocated to execute this complex maneuver effectively.

\subsection{Ablation Studies}
\textbf{Scene-aware Characterization.}
To validate the efficacy of our scene complexity characterization, we remove the scene-deviation, speed-deviation, and path-deviation tokens individually or jointly and compare the variants with the full model. As shown in Tab.~\ref{tab:ablation}, the full model achieves the highest DS and SR with the lowest average budget. Removing any deviation token impairs complexity estimation and budget allocation, leading to higher budgets and degraded performance. Removing the scene-deviation token only slightly increases the budget to 54.37\%, likely because visual tokens still retain scene-level information through prefix-layer fusion. In contrast, removing the speed-deviation or path-deviation token increases the budget to 71.89\% and 66.70\%, respectively, indicating that explicit historical decision uncertainty is critical for complexity assessment. When all deviation tokens are removed, the model requires the highest budget of 73.59\% while achieving lower DS and SR, further highlighting the complementary roles of three feature tokens encoding different deviations.
\begin{table}[H]
  \centering
  \setlength{\tabcolsep}{2pt}
  \caption{Ablation study on the closed-loop benchmark.}
  \label{tab:ablation}
  \begin{tabular*}{\linewidth}{@{\extracolsep{\fill}}cccc|ccccc@{}}
    \toprule
    Budget & Scene & Speed & Path & DS$\uparrow$ & SR(\%)$\uparrow$ & Efficiency$\uparrow$ & Comfortness$\uparrow$ & Avg. $B$ (\%)$\downarrow$ \\
    \midrule
    $\checkmark$ & $\checkmark$ & $\checkmark$ & $\checkmark$ & \textbf{84.87} & \textbf{65.45} & \textbf{266.19} & \underline{54.36} & \textbf{44.64} \\
    \midrule
    $\times$ & $\checkmark$ & $\checkmark$ & $\checkmark$ & 83.81 & 62.55 & \underline{259.18} & 51.49 & 60.03 \\
    \midrule
    $\checkmark$ & $\checkmark$ & $\times$ & $\checkmark$ & \underline{84.60} & \underline{64.09} & 249.68 & 37.63 & 71.89 \\
    $\checkmark$ & $\checkmark$ & $\checkmark$ & $\times$ & 83.88 & 63.63 & 247.17 & 30.75 & 66.70 \\
    $\checkmark$ & $\checkmark$ & $\times$ & $\times$ & 84.12 & 63.18 & 241.10 & 34.24 & 68.05 \\
    $\checkmark$ & $\times$ & $\checkmark$ & $\checkmark$ & 83.37 & \underline{64.09} & 258.74 & \textbf{56.74} & \underline{54.37} \\
    $\checkmark$ & $\times$ & $\times$ & $\times$ & 84.04 & 63.18 & 254.06 & 35.35 & 73.59 \\
    \bottomrule
  \end{tabular*}
\end{table}
\textbf{Budget-conditioned Layer Routing.}
To demonstrate the superiority of our budget-conditioned mechanism, we compare our scene-adaptive allocation against models utilizing random budgets and fixed budgets. As shown in Tab.~\ref{tab:ablation}, random allocation activates 60\% of parameters on average yet yields performance inferior to our dynamic 45\% activation.
Furthermore, with an average activation of merely 45\%, our method achieves a Success Rate falling between the 70\% and 100\% fixed-activation baselines as shown in Tab.~\ref{tab:bench2drive}. This shows that our scene-adaptive approach effectively reduces computational overhead in routine scenes while preserving essential capacity for complex scenarios.

\subsection{Generalization and Safety-Critical Analysis}
\textbf{Generalization.} Our primary experiments apply the Scene-Adaptive framework to SimLingo due to its strong closed-loop performance and efficiency. To evaluate generalization across VLA backbones, we further integrate SA into ORION, yielding ORION+SA. Unlike SimLingo, ORION adopts a Vicuna-v1.5 7B language backbone and a multi-view perception architecture.

As shown in Tab.~\ref{tab:FLOPS} and Tab.~\ref{tab:bench2drive}, ORION+SA reduces TFLOPs by 47.4\% and inference latency by 45.4\%, while maintaining competitive driving performance and outperforming DriveMoE. Its remaining gap from full ORION mainly stems from route deviations rather than safety-critical failures, as discussed in the next section. Moreover, with only about 53\% average activation, ORION+SA achieves DS and SR between those of the 65\% and 100\% fixed-activation baselines. These results demonstrate that SA generalizes to a substantially different VLA backbone, reducing computation in routine scenes while preserving sufficient capacity for complex ones.

\vspace{0.5em}
\noindent
\begin{minipage}[t]{0.53\textwidth}
  \vspace{0pt}
  \textbf{Safety-Critical Analysis.}
  Safety is critical for autonomous driving. To examine whether SA affects the
  safety-related capabilities of the original models, we analyze their failed
  Bench2Drive routes, focusing particularly on collision-related failures. We
  group failures into four categories and report the number of affected
  routes. Because a route may contain multiple violations, the category totals can exceed the total number of failed routes. Route Compliance (RC) mainly reflects route deviations, whereas Stability 
\end{minipage}\hfill
\begin{minipage}[t]{0.44\textwidth}
  \vspace{0pt}
  \centering
  \captionof{table}{Failed routes affected by each violation
  category.}
  \label{tab:safety_analysis}
  \small
  \setlength{\tabcolsep}{2pt}
  \renewcommand{\arraystretch}{1.12}
  \begin{tabular}{@{}l|cccc@{}}
    \toprule
    Category & SimLingo & \shortstack{+SA} &
    ORION & \shortstack{+SA} \\
    \midrule
    Collision            & 55.00 & 52.00 & 79.00 & 81.00 \\
    RC     & 13.00 & 19.00 & 26.00 & 37.00 \\
    S \& P & 17.00 & 13.00 & 12.00 & 10.00 \\
    TR         & 10.00 & 11.00 & 15.00 & 18.00 \\
    \bottomrule
  \end{tabular}
\end{minipage}
and Progress (S\&P) and Traffic Rule (TR) capture route completion and traffic-rule compliance.

As shown in Tab.~\ref{tab:safety_analysis}, collision-related failures do not
increase noticeably for either SA model and even decrease for SimLingo+SA,
indicating that SA does not introduce a substantial increase in observed collision-related failures. The main
increase occurs in Route Compliance, which primarily reflects route-following
stability rather than direct safety failures; the other categories remain
comparable or improve slightly. Thus, despite the small decrease in overall
driving performance, SA largely preserves the safety-critical driving
capabilities of the original models.

\section{Conclusions}
\label{sec:conclusions}
This study presents Scene-Adaptive VLA, an efficient framework designed to mitigate the substantial computational overhead and energy consumption of VLA models in autonomous driving. Addressing the underexplored area of structural adaptation, the proposed methodology implements a dynamic budget-conditioned layer routing mechanism that conditionally activates model parameters based on the real-time scene. By quantifying frame-level temporal scene variations and decision deviations, the framework intelligently allocates computational budgets and selects optimal layer combinations through an architecture optimized end-to-end. Evaluations on the Bench2Drive benchmark demonstrate that when adapted to a SOTA VLA driving model, Scene-Adaptive VLA reduces average inference-time computational overhead by 53.5\% with negligible degradation in driving and language capabilities.

\textbf{Limitations and Future Work:}
Although our framework is theoretically applicable to any dense VLA driving model, broader validation was limited by available computational resources. Future work will evaluate SA on more diverse VLA backbones, incorporate target-route deviation and auxiliary route-following loss to reduce the performance gap, and explore joint sequential and structural adaptation for further efficiency improvements.

\section{Acknowledgement}
This work was supported in part by the National Key Research and Development Program of China under Grant 2023YFB4301900, in part by the Natural Science Foundation of China under Grant 62272053, Grant 62671071, and Grant 62472048, in part by the Beijing Nova Program under Grant 202604841213, and in part by the International Academic Exchange Program for Doctoral Students of Beijing University of Posts and Telecommunications.

\bibliographystyle{unsrtnat}
\bibliography{refs}


\newpage
\appendix
\section{Details of Computational Overhead Evaluation}
\label{app:computational_overhead}
To specifically isolate and evaluate the impact of structural adaptation on inference computational overhead, we establish a fair and standardized configuration for input $N_{\text{in}}$ and output $N_{\text{out}}$ tokens across all evaluated models.
\begin{table}[htbp]
  \centering
  \small
  \caption{Token numbers across different models.}
  \label{tab:token_numbers}
  \begin{tabular}{l|ccc}
    \toprule
     & Visual Tokens & Prompt Tokens & Specific Tokens \\
    \midrule
    ORION & 512 & 10 & 16 \\
    SimLingo & 512 & 25 & 32 \\
    AutoVLA & 1536 & 200 & 0 \\
    DriveMoE & 512 & 10 & 5 \\
    \bottomrule
  \end{tabular}
\end{table}
Input Tokens ($N_{\text{in}}$): The input sequence comprises frame-level visual tokens, language prompt tokens, and model-specific auxiliary tokens, with precise quantities inferred from their respective original papers, as shown in Tab.~\ref{tab:token_numbers}.
The predominant factor influencing $N_{\text{in}}$ is the visual token count, which is directly determined by the number of frames input per decision. For instance, AutoVLA processes four consecutive frames per decision, inherently resulting in a significantly larger input token sequence compared to other models that typically rely on a two-frame average.
Output Tokens ($N_{\text{out}}$): it denotes the number of autoregressively generated output tokens, including language and action tokens where applicable. For models that predict actions non-autoregressively, $N_{\text{out}}$ is set to 1, corresponding to a single action prediction step. For autoregressive models, we use the model-specific average number of output tokens per step on the Bench2Drive benchmark, thereby accounting for variations in output length across models.

To ensure a fair comparison for models possessing dynamic inference capabilities, we evaluate them under their minimal configurations. AutoVLA is evaluated in fast-thinking mode, bypassing textual reasoning and autoregressively generating solely action tokens. Similarly, DriveMoE dynamically selects input views, allocating two consecutive frames per view. For our evaluation, we conservatively configure DriveMoE to utilize a single-view input.

\section{Bench2Drive Multi-Ability Results}
\label{app:multi-ability}
Multi-ability testing utilizes short, scenario-specific routes to isolate individual driving skills, effectively reducing metric variance and revealing specific system strengths.
\begin{table}[h]
    \centering
    \caption{Multi-ability evaluation on Bench2Drive.}
    \begin{tabular}{l|ccccc|c}
        \toprule
        \multirow{2}{*}{\textbf{Method}} & \multicolumn{5}{c|}{\textbf{Ability (\%) $\uparrow$}} & \multirow{2}{*}{\textbf{Mean}} \\
        \cmidrule{2-6}
        & Merging & Overtaking & Emergency Brake & Give Way & Traffic Sign & \\
        \midrule
        UniAD-Base & 14.10 & 17.78  & 21.67 & 10.00 & 14.21  & 15.55 \\
        VAD        & 8.11  & 24.44  & 18.64 & 20.00 & 19.15  & 18.07 \\
        TCP-traj   & 8.89  & 24.29  & 51.67 & 40.00 & 46.28  & 34.22 \\
        Orion      & 25.00 & \textbf{71.11}  & 78.33 & 30.00 & 69.15  & 54.72 \\
        DriveMoE   & 34.67 & 40.00  & 65.45 & 40.00 & 59.44  & 47.91 \\
        simlingo   & \textbf{54.01} & 57.04  & \textbf{88.33} & \textbf{53.33} & \textbf{82.45}  & \textbf{67.03} \\
        SimLing+SA & \underline{47.50} & \underline{68.88} & \underline{80.00}   & \underline{50.00}   & \underline{80.52} & \underline{65.38} \\
        \bottomrule
    \end{tabular}
    \label{tab:multi-ability}
\end{table}
It assesses performance across distinct interactive scenarios, including merging, overtaking, Emergency Brake, Give Way, and Traffic Sign, and provides a comprehensive and decoupled evaluation of a model's driving capabilities under varied conditions.
Although our method decreases the multi-ability mean score of the original model by 1.65\%, it significantly enhances its overtaking capability by 11.84\%, as shown in Tab.~\ref{tab:multi-ability}.

\section{Training Details}
\label{app:training_details}
We present the pseudocode for the two-stage training pipeline. The notation follows Section \ref{sec:Scene-Adaptive-VLA}.

\subsection{Stage 1: Scene-Aware Representation and Layer Routing Initialization}

Stage~1 jointly initializes the scene-aware encoders, the routing scheduler, the budget encoder, the layer routing query, and the LoRA-tuned prefix layers via supervised fine-tuning with randomized budgets. The budget scheduler $f_{\text{budget}}$ and budget query $\mathbf{q}_b$ remain frozen. 
Algorithm~\ref{alg:stage1} details the procedure, with the Straight-Through Gumbel-Softmax top-$k$ estimator highlighted.

\begin{algorithm}[H]
\caption{Stage 1: Scene-Aware Representation and Layer Routing Initialization}
\label{alg:stage1}
\linespread{1.25}\selectfont
\textbf{Input:} Training dataset $\mathcal{D} = \{(\mathbf{e}_I, p_{\text{task}}, \mathbf{e}_{\text{nav}}, v, \mathbf{W}_p^{\text{gt}}, \mathbf{W}_s^{\text{gt}}, l_{1:T})\}$, Prefix layers $L_{\text{pre}}$, Sub-layers $L_{\text{sub}}$, Budget lower bound $b_{\min}$, Temperature $\tau$, Loss weights $\lambda_{\text{wp}}, \lambda_{\text{lang}}, \lambda_{\text{div}}, \lambda_{\text{bal}}$ \\
\textbf{Output:} Optimized parameters for scene-adaptive routing $\Theta_{\text{train}}$
\begin{algorithmic}[1]
\STATE Initialize trainable parameters $\Theta_{\text{train}} \leftarrow \{\mathbf{q}_{lr}, f_{\text{routing}}, f_{\text{enc}}, \text{scene encoders}, \text{prefix LoRA}\}$
\STATE Freeze budget scheduler and query $\Theta_{\text{frozen}} \leftarrow \{f_{\text{budget}}, \mathbf{q}_b\}$
\FOR{each training step}
    \STATE Sample batch from $\mathcal{D}$
    \STATE Extract scene-aware tokens $\mathbf{e}_s, \mathbf{e}_{pw}, \mathbf{e}_{sw}$ via Eq.~\eqref{eq:cross_attn}--\eqref{eq:speed_dev_token}
    \STATE Form input sequence $I_{\text{llm}} \leftarrow [p_{\text{global}}, \mathbf{q}_p, \mathbf{q}_s, \mathbf{e}_s, \mathbf{e}_{pw}, \mathbf{e}_{sw}, \mathbf{q}_b, \mathbf{q}_{lr}]$
    \STATE Extract budget feature $\mathbf{h}_{q_b} \leftarrow \text{PrefixLayers}_{1 \to L_{\text{pre}}/2}(I_{\text{llm}})$ 
    \STATE Sample random budget $b \sim \mathcal{U}[b_{\min}, 1]$
    \STATE Quantize active layer count $k \leftarrow \max(0, \lfloor (L_{\text{pre}} + L_{\text{sub}}) \cdot b \rfloor - L_{\text{pre}})$
    \STATE Encode budget $\mathbf{e}_b \leftarrow f_{\text{enc}}(\text{Sinusoidal}(b))$ and replace $\mathbf{q}_b$ in intermediate features
    \STATE \text{Updated features} $\mathbf{h}_{q_{lr}} \leftarrow \text{PrefixLayers}_{L_{\text{pre}}/2+1 \to L_{\text{pre}}}$
    \STATE Compute routing logits $\boldsymbol{\pi}_{\text{layer}} \leftarrow f_{\text{routing}}([\mathbf{h}_{q_{lr}}, \mathbf{e}_b]) \in \mathbb{R}^{L_{\text{sub}}}$
    \STATE Straight-Through Gumbel-Softmax Top-$k$ estimation:
    \STATE \quad $\mathbf{s} \leftarrow \text{softmax}\big((\boldsymbol{\pi}_{\text{layer}} + \mathbf{g}) / \tau\big),\mathbf{g} \sim \text{Gumbel}(0,1)$
    \STATE \quad $\mathbf{M}_{\text{hard}} \leftarrow \text{OneHotTopK}(\mathbf{s})$
    \STATE \quad Mask $\mathbf{M} \leftarrow \texttt{stopgrad}(\mathbf{M}_{\text{hard}} - \mathbf{s}) + \mathbf{s}$
    \STATE Route and decode $\mathbf{W}_p, \mathbf{W}_s, l_{1:T} \leftarrow \text{SubLayers}(\text{features} \mid \mathbf{M})$ 
    \STATE Compute $\mathcal{L}_{\text{wp}}, \mathcal{L}_{\text{lang}}$ via Eq.~\eqref{eq:loss_wp_lang} and $\mathcal{L}_{\text{div}}, \mathcal{L}_{\text{bal}}$ via Eq.~\eqref{eq:loss_div_bal}
    \STATE $\mathcal{L}_{\text{stage1}} \leftarrow \lambda_{\text{wp}} \mathcal{L}_{\text{wp}} + \lambda_{\text{lang}} \mathcal{L}_{\text{lang}} + \lambda_{\text{div}}\mathcal{L}_{\text{div}} + \lambda_{\text{bal}}\mathcal{L}_{\text{bal}}$
    \STATE Optimize $\Theta_{\text{train}}$ via $\nabla_{\Theta_{\text{train}}} \mathcal{L}_{\text{stage1}}$
\ENDFOR
\end{algorithmic}
\end{algorithm}

\noindent The diversity regularizer uses max-pooled scene features $\mathbf{z}_i$ aggregated from visual tokens, prompt tokens, and the budget token:
$\mathbf{z}_i = \max\text{pool}\big(\{\mathbf{e}_{I}^{i,j}\}_{j=1}^{N_c} \cup \{\mathbf{e}_{\text{prompt}}^{i,t}\}_{t} \cup \{\mathbf{e}_b^{i}\}\big)$.
The cosine similarity between routing masks $\mathbf{M}_i$ is contrasted with that of scene features $\mathbf{z}_i$:
$\text{sim}(\mathbf{M}_i, \mathbf{M}_j) = \frac{\mathbf{M}_i \cdot \mathbf{M}_j}{\|\mathbf{M}_i\|_2 \|\mathbf{M}_j\|_2}$, and analogously for $\text{sim}(\mathbf{z}_i, \mathbf{z}_j)$.

\subsection{Stage 2: GRPO with Budget Allocation Policy}

Stage~2 freezes the routing pathway and optimizes the budget allocation policy $\pi_\theta$ via Group Relative Policy Optimization (GRPO). Crucially, since the budget query feature $\mathbf{h}_{q_b}$ is extracted midway through the prefix layers, we explicitly unfreeze only the first half of the prefix layers ($1$ to $L_{\text{pre}}/2$) alongside the budget scheduler, while freezing the latter half to preserve the routing stability established in Stage 1. The policy $\pi_\theta(a \mid \mathbf{h}_{q_b}) = \text{Categorical}\big(\text{softmax}(f_{\text{budget}}(\mathbf{h}_{q_b}))\big)$ maps the intermediate budget query representation to a distribution over $K$ discrete budget actions. Algorithm~\ref{alg:stage2} presents the procedure.

\begin{algorithm}[H]
\caption{Stage 2: Reinforcement Fine-Tuning for Budget Allocation}
\label{alg:stage2}
\linespread{1.25}\selectfont
\textbf{Input:} Training dataset $\mathcal{D}$, Initial policy $\pi_{\theta}$ from Stage 1, Group size $G$, Update epochs $E$, Budget mapping $b(a)$, Reward weights $w_d, w_b, w_a$, Clipping range $\varepsilon_c$, KL weight $\beta$, Anchor weight $\beta_{\text{anchor}}$ \\
\textbf{Output:} Optimized budget allocation policy $\pi_{\theta}$
\begin{algorithmic}[1]
\STATE Unfreeze policy parameters $\theta \leftarrow \{f_{\text{budget}}, \mathbf{q}_b, \text{scene encoders}, \text{prefix LoRA}_{1 \to L_{\text{pre}}/2}\}$
\STATE Freeze routing parameters $\Theta_{\text{frozen}} \leftarrow \{f_{\text{routing}}, f_{\text{enc}}, \mathbf{q}_{lr}, \text{base LLM}, \text{prefix LoRA}_{L_{\text{pre}}/2+1 \to L_{\text{pre}}}\}$
\STATE Store reference anchor parameters $\bar{\boldsymbol{\phi}} \leftarrow \boldsymbol{\phi}_{\text{Stage1}}$ and initial policy $\pi_{\text{old}} \leftarrow \pi_\theta$

\FOR{each training step}
    \STATE Sample batch from $\mathcal{D}$ and extract budget query features $\{\mathbf{h}_{q_b}^{(i)}\}_{i=1}^B$
    \FOR{each sample $i \in \{1, \dots, B\}$}
        \STATE Sample $G$ actions $\{a_{i,g}\}_{g=1}^G \sim \text{Categorical}\big(\text{softmax}(f_{\text{budget}}(\mathbf{h}_{q_b}^{(i)}))\big)$ under $\pi_{\text{old}}$
        \STATE Execute forward pass for each $a_{i,g}$ to obtain driving loss $\mathcal{L}_{i,g}$
        \STATE Compute reward $r_{i,g}$ and relative advantage $A_{i,g}$ via Eq.~\eqref{eq:reward_advantage}
    \ENDFOR
    
    \FOR{epoch $= 1$ to $E$}
        \STATE Evaluate importance sampling ratio $\rho_{i,g}(\theta) \leftarrow \frac{\pi_\theta(a_{i,g} \mid \mathbf{h}_{q_b}^{(i)})}{\pi_{\text{old}}(a_{i,g} \mid \mathbf{h}_{q_b}^{(i)})}$
        \STATE Compute policy objective $\mathcal{L}_{\text{policy}}$ via Eq.~\eqref{eq:loss_policy}
        \STATE Compute anchor loss $\mathcal{L}_{\text{anchor}} \leftarrow \frac{1}{|\boldsymbol{\phi}|}\sum_j (\phi_j - \bar{\phi}_j)^2$
        \STATE Evaluate task driving loss $\mathcal{L}_{\text{driving}} \leftarrow \lambda_{\text{wp}} \mathcal{L}_{\text{wp}} + \lambda_{\text{lang}} \mathcal{L}_{\text{lang}}$ under current $\pi_\theta$
        \STATE $\mathcal{L}_{\text{stage2}} \leftarrow \mathcal{L}_{\text{policy}} + \beta_{\text{anchor}} \mathcal{L}_{\text{anchor}} + w_{\text{driving}}\mathcal{L}_{\text{driving}}$
        \STATE Optimize $\theta$ via $\nabla_{\theta} \mathcal{L}_{\text{stage2}}$
    \ENDFOR
    \STATE Synchronize reference policy $\pi_{\text{old}} \leftarrow \pi_\theta$
\ENDFOR
\end{algorithmic}
\end{algorithm}

\noindent The direct driving loss weight follows a linear warmup, preventing early-stage policy collapse. The anchor parameters $\bar{\boldsymbol{\phi}}$ are snapshots of the corresponding Stage~1 trained weights and are never updated during Stage~2. The budget action space consists of $K = 10$ discrete actions, where action $a \in \{0,\dots,9\}$ maps to budget $b(a) = \max(0.1 \cdot (a+1),\, b_{\min})$.
To encourage exploration in early Stage~2, we optionally apply temperature annealing and $\epsilon$-greedy sampling when drawing actions from the old policy.

\newpage
\section*{NeurIPS Paper Checklist}

\begin{enumerate}

\item {\bf Claims}
    \item[] Question: Do the main claims made in the abstract and introduction accurately reflect the paper's contributions and scope?
    \item[] Answer: \answerYes{} 
    \item[] Justification: The claims made in the abstract and introduction are empirically verified in Section \ref{sec:experiments}, in particular Table \ref{tab:bench2drive}.

\item {\bf Limitations}
    \item[] Question: Does the paper discuss the limitations of the work performed by the authors?
    \item[] Answer: \answerYes{} 
    \item[] Justification: Our limitations are discussed in Section \ref{sec:conclusions}.

\item {\bf Theory assumptions and proofs}
    \item[] Question: For each theoretical result, does the paper provide the full set of assumptions and a complete (and correct) proof?
    \item[] Answer: \answerNA{} 
    \item[] Justification: The paper does not include theoretical results.

    \item {\bf Experimental result reproducibility}
    \item[] Question: Does the paper fully disclose all the information needed to reproduce the main experimental results of the paper to the extent that it affects the main claims and/or conclusions of the paper (regardless of whether the code and data are provided or not)?
    \item[] Answer: \answerYes{} 
    \item[] Justification: We disclose all information needed for reproduction in Section \ref{sec:experiments} and Appendix \ref{app:training_details}.

\item {\bf Open access to data and code}
    \item[] Question: Does the paper provide open access to the data and code, with sufficient instructions to faithfully reproduce the main experimental results, as described in supplemental material?
    \item[] Answer: \answerYes{} 
    \item[] Justification: We will provide data, code, and instructions for the final paper.

\item {\bf Experimental setting/details}
    \item[] Question: Does the paper specify all the training and test details (e.g., data splits, hyperparameters, how they were chosen, type of optimizer) necessary to understand the results?
    \item[] Answer: \answerYes{} 
    \item[] Justification: We provide experimental details in Section \ref{sec:experiments} and Appendix \ref{app:training_details}.

\item {\bf Experiment statistical significance}
    \item[] Question: Does the paper report error bars suitably and correctly defined or other appropriate information about the statistical significance of the experiments?
    \item[] Answer: \answerYes{} 
    \item[] Justification: We conduct our experiments with 3 random seeds and report the mean of the results as detailed in Section \ref{sec:experiments}.

\item {\bf Experiments compute resources}
    \item[] Question: For each experiment, does the paper provide sufficient information on the computer resources (type of compute workers, memory, time of execution) needed to reproduce the experiments?
    \item[] Answer: \answerYes{} 
    \item[] Justification: We illustrate the computer resources we use in Section \ref{sec:experiments}.
    
\item {\bf Code of ethics}
    \item[] Question: Does the research conducted in the paper conform, in every respect, with the NeurIPS Code of Ethics \url{https://neurips.cc/public/EthicsGuidelines}?
    \item[] Answer: \answerYes{} 
    \item[] Justification: We conducted the research in the paper in conformance with the NeurIPS Code of Ethics.
    \item[] Guidelines:
    \begin{itemize}
        \item The answer \answerNA{} means that the authors have not reviewed the NeurIPS Code of Ethics.
        \item If the authors answer \answerNo{}, they should explain the special circumstances that require a deviation from the Code of Ethics.
        \item The authors should make sure to preserve anonymity (e.g., if there is a special consideration due to laws or regulations in their jurisdiction).
    \end{itemize}

\item {\bf Broader impacts}
    \item[] Question: Does the paper discuss both potential positive societal impacts and negative societal impacts of the work performed?
    \item[] Answer: \answerYes{} 
    \item[] Justification: Our work contributes to the advancement of VLA-based autonomous driving technology as introduced in Section \ref{sec:introduction}.
    \item[] Guidelines:
    \begin{itemize}
        \item The answer \answerNA{} means that there is no societal impact of the work performed.
        \item If the authors answer \answerNA{} or \answerNo{}, they should explain why their work has no societal impact or why the paper does not address societal impact.
        \item Examples of negative societal impacts include potential malicious or unintended uses (e.g., disinformation, generating fake profiles, surveillance), fairness considerations (e.g., deployment of technologies that could make decisions that unfairly impact specific groups), privacy considerations, and security considerations.
        \item The conference expects that many papers will be foundational research and not tied to particular applications, let alone deployments. However, if there is a direct path to any negative applications, the authors should point it out. For example, it is legitimate to point out that an improvement in the quality of generative models could be used to generate Deepfakes for disinformation. On the other hand, it is not needed to point out that a generic algorithm for optimizing neural networks could enable people to train models that generate Deepfakes faster.
        \item The authors should consider possible harms that could arise when the technology is being used as intended and functioning correctly, harms that could arise when the technology is being used as intended but gives incorrect results, and harms following from (intentional or unintentional) misuse of the technology.
        \item If there are negative societal impacts, the authors could also discuss possible mitigation strategies (e.g., gated release of models, providing defenses in addition to attacks, mechanisms for monitoring misuse, mechanisms to monitor how a system learns from feedback over time, improving the efficiency and accessibility of ML).
    \end{itemize}
    
\item {\bf Safeguards}
    \item[] Question: Does the paper describe safeguards that have been put in place for responsible release of data or models that have a high risk for misuse (e.g., pre-trained language models, image generators, or scraped datasets)?
    \item[] Answer: \answerNo{} 
    \item[] Justification: Our paper does not release data or models that have a high risk for misuse.

\item {\bf Licenses for existing assets}
    \item[] Question: Are the creators or original owners of assets (e.g., code, data, models), used in the paper, properly credited and are the license and terms of use explicitly mentioned and properly respected?
    \item[] Answer: \answerYes{} 
    \item[] Justification: We use publicly available datasets and models with proper citations and respect for their licenses.

\item {\bf New assets}
    \item[] Question: Are new assets introduced in the paper well documented and is the documentation provided alongside the assets?
    \item[] Answer: \answerYes{} 
    \item[] Justification: We will release our codebase along with detailed README files for our work.

\item {\bf Crowdsourcing and research with human subjects}
    \item[] Question: For crowdsourcing experiments and research with human subjects, does the paper include the full text of instructions given to participants and screenshots, if applicable, as well as details about compensation (if any)? 
    \item[] Answer: \answerNA{} 
    \item[] Justification: This work does not involve crowdsourcing nor research with human subjects.

\item {\bf Institutional review board (IRB) approvals or equivalent for research with human subjects}
    \item[] Question: Does the paper describe potential risks incurred by study participants, whether such risks were disclosed to the subjects, and whether Institutional Review Board (IRB) approvals (or an equivalent approval/review based on the requirements of your country or institution) were obtained?
    \item[] Answer: \answerNA{} 
    \item[] Justification: This work does not involve crowdsourcing nor research with human subjects.

\item {\bf Declaration of LLM usage}
    \item[] Question: Does the paper describe the usage of LLMs if it is an important, original, or non-standard component of the core methods in this research? Note that if the LLM is used only for writing, editing, or formatting purposes and does \emph{not} impact the core methodology, scientific rigor, or originality of the research, declaration is not required.

    \item[] Answer: \answerNA{} 
    \item[] Justification: The core method development in this research does not involve LLMs as an important, original, or non-standard component.

\end{enumerate}

\end{document}